\documentclass[journal]{IEEEtran}

\usepackage{xcolor,soul,framed} 
\usepackage[printonlyused]{acronym}
\usepackage{float}
\usepackage{subfig}
\usepackage{algorithm}
\usepackage{algpseudocode}
\usepackage{bbm}
\usepackage{comment}
\makeatletter
\newcommand\fs@norules{\def\@fs@cfont{\bfseries}\let\@fs@capt\floatc@ruled
  \def\@fs@pre{}%
  \def\@fs@post{}%
  \def\@fs@mid{\kern3pt}%
  \let\@fs@iftopcapt\iftrue}
\makeatother
\colorlet{shadecolor}{yellow}
\usepackage[pdftex]{graphicx}
\graphicspath{{../pdf/}{../jpeg/}}
\DeclareGraphicsExtensions{.pdf,.jpeg,.png}

\usepackage[cmex10]{amsmath}
\usepackage{array}
\usepackage{mdwmath}
\usepackage{mdwtab}
\usepackage{eqparbox}
\usepackage{url}
\usepackage{amsfonts}
\usepackage{cleveref}
\usepackage{subfig}
\usepackage[scr=rsfs]{mathalpha}
\usepackage{makecell}
\usepackage{float} 
\algrenewcommand\algorithmicrequire{\textbf{Input:}}
\algrenewcommand\algorithmicensure{\textbf{Output:}}
\crefalias{subequation}{equation} 
\crefalias{eqnarray}{equation} 
\crefformat{pluraleq}{(#2#1#3)}

\usepackage{bm}

\newcommand{\cl}{\textcolor{black}}

\DeclareUnicodeCharacter{2212}{-}
\begin{document}
\bstctlcite{IEEEexample:BSTcontrol}
    \title{Spectral Cross-Domain Neural Network with Soft-adaptive Threshold Spectral Enhancement}
\author{Che Liu, Sibo Cheng, Weiping Ding, Senior Member, IEEE, and Rossella Arcucci
      
\thanks{Corresponding author: Sibo Cheng (sibo.cheng@imperial.ac.uk)}
\thanks{Che Liu and Rossella Arcucci are with Department of Earth Science and Engineering, Imperial College London, SW7 2AZ, UK. }
\thanks{Che Liu is also with Data Science Institute, Department of computing, Imperial College London, SW7 2AZ, UK. }
\thanks{Sibo Cheng is with Data Science Institute, Department of computing, Imperial College London, SW7 2AZ, UK. }
\thanks{Weiping Ding is with School of Information Science and Technology, Nantong University, Nantong 226019, China. }
}

\markboth{IEEE TRANSACTIONS ON Neural Networks and Learning Systems, VOL. XX, NO. XX, XXXX
2023}{Che {\textit{et al.}}: Spectral Cross-Domain Neural Network with Soft-adaptive Threshold Spectral
Enhancement}

\maketitle

\begin{abstract}
Electrocardiography (ECG) signals can be considered as multi-variable time-series. The state-of-the-art ECG data classification approaches, based on either feature engineering or deep learning techniques, treat separately spectral and time domains in machine learning systems. No spectral-time domain communication mechanism inside the classifier model can be found in current approaches, leading to difficulties in identifying complex ECG forms. In this paper, we proposed a novel deep learning model named Spectral Cross-domain neural network (SCDNN) with a new block called Soft-adaptive threshold spectral
enhancement (SATSE), to simultaneously reveal the key information embedded in spectral and time domains inside the neural network. More precisely, the domain-cross information is captured by a general Convolutional neural
network (CNN) backbone, and different information sources are merged by a self-adaptive mechanism to mine the connection between time and spectral domains. In SATSE, the knowledge from time and spectral domains is extracted via the Fast Fourier Transformation (FFT) with soft trainable thresholds in modified Sigmoid functions. The proposed SCDNN is tested with several classification tasks implemented on the public ECG databases \textit{PTB-XL} and \textit{CPSC2018}. SCDNN 
outperforms the state-of-the-art approaches with a low computational cost regarding a variety of metrics in all classification tasks on both databases, by finding appropriate domains from the infinite spectral mapping. The convergence of the trainable thresholds in the spectral domain is also numerically investigated in this paper. The robust performance of SCDNN provides a new perspective to exploit knowledge across deep learning models from time and spectral domains. The code repository can be found: https://github.com/DL-WG/SCDNN-TS
\end{abstract}


\IEEEpeerreviewmaketitle


\section{Introduction}\label{sec3}

Time Series has been widely studied in research fields such as physics, finance, medical, and nature language processing~\cite{hamilton2020time}.
Due to the time dependency~\cite{bagnall2017great}, classifying  TS is different from traditional classification tasks involving image or sequence classifications without time dependencies~\cite{ewusie2020methods}. The performance of traditional classification tasks often relies on the number of samples with proper labels~\cite{wen2020time}. In fact, well-labelled TS is often out of reach in real applications as stated in~\cite{bagnall2017great}. Moreover, the issue of imbalanced datasets commonly exists in medical Time Series data~\cite{wen2020time}. As pointed out by~\cite{ewusie2020methods}, another critical challenge of time series classification involves the model generalization. It is found that different models are often required for distinct classification tasks even with the same input Time Series data~\cite{ngiam2011multimodal}. 

The state-of-the-art time series classification approaches are either based on handcrafted feature engineering~\cite{bagnall2017great,zhao2005ecg,kallas2012multi,li2016ecg,kropf2017ecg} or deep learning methods~\cite{ ismail2019deep,mostayed2018classification, yang2017localization,liu2018automatic,kachuee2018ecg}.  The former can be divided into three main categories, namely statistical feature extractions~\cite{anderson2011statistical,luo2020position,luo2021novel}, entropy-based methods~\cite{fulcher2014highly} and frequency-based methods~\cite{lines2015time}. After preprocessing~\cite{faloutsos1994fast}, traditional machine learning classifiers such as \ac{GBDT}, \ac{SVM} and \ac{RF} are employed to accomplish the classification tasks using handcrafted features as model input. However, the performance of FE-based methods is extremely sensitive to the quality of feature extractions on a case-to-case basis~\cite{halliday1995framework}. Furthermore, distinct FE methods are often required for different time series classification tasks even for the same dataset~\cite{box2015time}. Thus, there is an insurmountable obstacle to extending specific feature engineering methods to general classification tasks.

Such challenges can be found in medical time series classification as stated in the work of~\cite{sternickel2002automatic,chuah2007ecg}.
\ac{ECG} is a specific type of medical TS that describes the heartbeats of 12 different leads to detect various aspects of heart health~\cite{van1997clinical}. In clinical applications, ECG data are widely used to diagnose cardiovascular diseases, such as myocardial infarction(MI), hypertrophy(HYP), and Conduction Disturbance(CD)~\cite{ashley2001evidence}. A standard ECG TS consists of three waves, known as P-wave, QRS-complex and T-wave respectively~\cite{hurst1998naming}. 
As recognised in numerous researches~\cite{rautaharju2009aha,de1998prognostic,maron2014assessment}, various cardiovascular diseases will impact these three waves, resulting in unrecognisable signals (also called repolarization in medical science). Thus identifying these abnormal signals is crucial for clinical diagnosis.
In fact, successful \ac{ECG} classification methods not only improve the accuracy of cardiology diagnosis, but also enable the possibility of monitoring the state of human health state using wearable devices.

Classical feature engineering-based time series classification approaches extract the features relying on different waveforms of 12-lead \ac{ECG}, mostly focusing on the P-QRS-T wave~\cite{hurst1998naming} and RR-interval~~\cite{brown1993important}. feature engineering is carried out to extract statistical, energy-based and frequency-based features which are used  to classify \ac{ECG} Time Series via traditional machine learning classifiers~\cite{zhao2005ecg,kallas2012multi,li2016ecg,kropf2017ecg}. In feature engineering based methods, the quality of the features and thus the classification performance is sensitive to the algorithms selected for feature computation~\cite{ma2021image}. Therefore, specific feature engineering algorithms often need to be designed for different classification tasks.

In recent years, much research attention has been given to applying deep learning approaches for \ac{ECG} classification, and more generally, for time series classification tasks. For example,  \acp{CNN} have been widely employed to classify 12-lead \ac{ECG} signals~\cite{mostayed2018classification, yang2017localization,liu2018automatic,kachuee2018ecg}, improving the accuracy of disease diagnosis in comparison to traditional approaches.
The works of~\cite{hou2019lstm,gao2019effective} have used \ac{LSTM}~\cite{hochreiter1997long}, a variant of \ac{RNN}, to classify \ac{ECG} pattern with imbalanced data.
A wavelet layer before \ac{LSTM} has been added in~\cite{ yildirim2018novel} to merge spectral domain knowledge into the \ac{NN}. In their work, the spectral operation has been performed only on the input \ac{ECG} Time Series, instead of processing deep knowledge inside the \ac{NN}.

Recent works of~\cite{wang2020deep,prabhakararao2021multi} have used multi-scale deep convolutional neural networks with ensemble learning to detect heart arrhythmia from 12-lead \ac{ECG}.  MLFB-Net~\cite{zhang2021mlbf} with \ac{CNN} concatenated bidirectional GRU~\cite{dey2017gate} and attention mechanism~\cite{vaswani2017attention} has also been implemented for \ac{ECG} classification where each lead \ac{ECG} is treated individually. 
In addition, the work of~\cite{li2021bat} has tried the Transformer structure with an attention mechanism to capture latent and deep knowledge from 12-lead \ac{ECG} simultaneously. However, their works only consider the \ac{ECG} signals in time domain, and neglect the valuable information embedded in the spectral domain.

\cl{Recent researches, such as~\cite{wang2020deep,prabhakararao2021multi}, utilise multi-scale deep convolutional neural networks and ensemble learning for heart arrhythmia detection from 12-lead \ac{ECG}. Another methodology, the MLFB-Net~\cite{zhang2021mlbf}, employs \ac{CNN} coupled with a concatenated bidirectional Gated Recurrent Unit~\cite{dey2017gate} and attention mechanism~\cite{vaswani2017attention} for ECG classification.
\cite{li2021bat} has incorporated the Transformer structure with an attention mechanism to simultaneously extract latent and deeper knowledge from 12-lead ECG, although these studies primarily focus on time domain ECG signals and may overlook spectral domain information.
Our model is compared with various state-of-the-art methodologies. These include SR2-CF2~\cite{SR2-CF2}, a feature-based model for time series classification that utilises a genetic algorithm; EARLIEST~\cite{EARLIEST}, a reinforcement learning approach that outputs classification results via a trained policy network; TEASER~\cite{TEASER}, designed to handle varied-length time series early classification problems using subclassifiers; MDDNN~\cite{MDDNN}, a deep learning model that combines \ac{CNN} and \ac{LSTM} for time series classification; ETEeTSC~\cite{ETEeTSC}, a deep learning model that simultaneously optimises accuracy and earliness; SPN~\cite{SPN}, a deep reinforcement learning solution for time series classification combining a deep learning backbone with a trained policy network; SPNv2~\cite{huang2022snippet}, a sophisticated deep reinforcement learning framework, intra-snippet spatial correlations and inter-snippet temporal correlations are integrated into concealed ECG representations, and CKNA~\cite{huang2022novel}, employs the strategy of imposing constraints on specific diseases, aiming to enhance the accuracy of diagnosis and early detection outcomes. }

The knowledge representation on the spectral domain of images has been introduced in the work of~\cite{rippel2015spectral}. Following this idea,~\cite{yi2017syncspeccnn,han2018ssf,mou2019learning,alqudah2020aoct} have further explored the spectral domain representation of image data on \acp{CNN}.  Nevertheless, none of the aforementioned studies has established the links between the time domain and the spectral domain inside the \ac{NN}. In other words, the spectral information is either processed outside the neural network~\cite{rippel2015spectral,yi2017syncspeccnn} or inside the \ac{NN} but without connections to the time domain. The very recent works of~\cite{li2020fourier,pathak2022fourcastnet} have extended spectral domain knowledge to an infinite mapping space to approximate the proper space with \ac{FNO} for image prediction tasks. In~\cite{rao2021global}, \ac{FNO} has been utilized to mix multi-scale features in the spectral domain to reach a more general representation of images. However, in these works, the dimension of the spectral domain is fixed  and only low-frequency domain information is kept in the neural network. Hence, it is extremely challenging to obtain the proper spectral domain due to the pre-fixed threshold, and the high-frequency domain is ignored. The latter may lead to potential information loss due to the hard thresholds~\cite{tanter2000time}.\\

As discussed, although much effort has been dedicated to time series classification involved spectral knowledge, the noted algorithms suffer from the following limitations and challenges:

\begin{enumerate}
    \item  Current approaches either process the spectral information outside the deep learning model or treat time and spectral domains separately in the neural network. No communication between time and spectral domains inside the neural network has been performed;
    \item When filtering information in the spectral domain, pre-selected number of spectral modes are employed, leading to potential information loss.
\end{enumerate}

To overcome these bottlenecks, we propose a novel neural network structure, named Spectral Cross-domain neural network (SCDNN), that interacts the time domain and the spectral domain inside the \acp{NN}. In addition, we have developed a new soft self-adaptive threshold determination mechanism that is deployed in the \ac{SATSE} block of \ac{SCDNN}. More
precisely, this new mechanism enables \ac{SCDNN} to find the optimal number of spectral modes instead of using pre-fixed thresholds as implemented in existing approaches. Furthermore, the information loss of spectral filtering can be decreased thanks to the controllable soft threshold. To verify our model's performance and versatility, four diverse \ac{ECG} classification tasks are deployed. 
These tasks consist of 14 different cardiovascular diseases classifications on the public datasets $PTB-XL$ and $CPSC2018$. \ac{SCDNN} achieves substantial enhancement compared to existing time series classification models, in relation to almost all comparison metrics evaluated. 

In summary, the principle contributions of our work are listed below:
\begin{enumerate}
    \item We designed a novel neural network structure, named \ac{SCDNN}, to establish the communication of spectral and time domains inside \acp{NN}. More precisely, \ac{FFT} and \ac{IFFT} are added on the ResNet backbone after each Res block to learn the information in the spectral domain. 
    
    \item In the \ac{SATSE} block of \ac{SCDNN}, trainable soft thresholds in the spectral domain enable the proposed model to find optimal spectral modes and reduce the information loss compared to using pre-fixed thresholds.
    
    \item  \cl{The efficacy of the proposed SCDNN and SATSE has been rigorously assessed using two publicly accessible, large-scale datasets, namely $PTB-XL$ and $CPSC2018$. All results attest to the superior performance of the SCDNN in cardiovascular disease classification across all four evaluation metrics.}
\end{enumerate}

\begin{figure*}[htp!]
\centering
\includegraphics[width=0.9\textwidth]{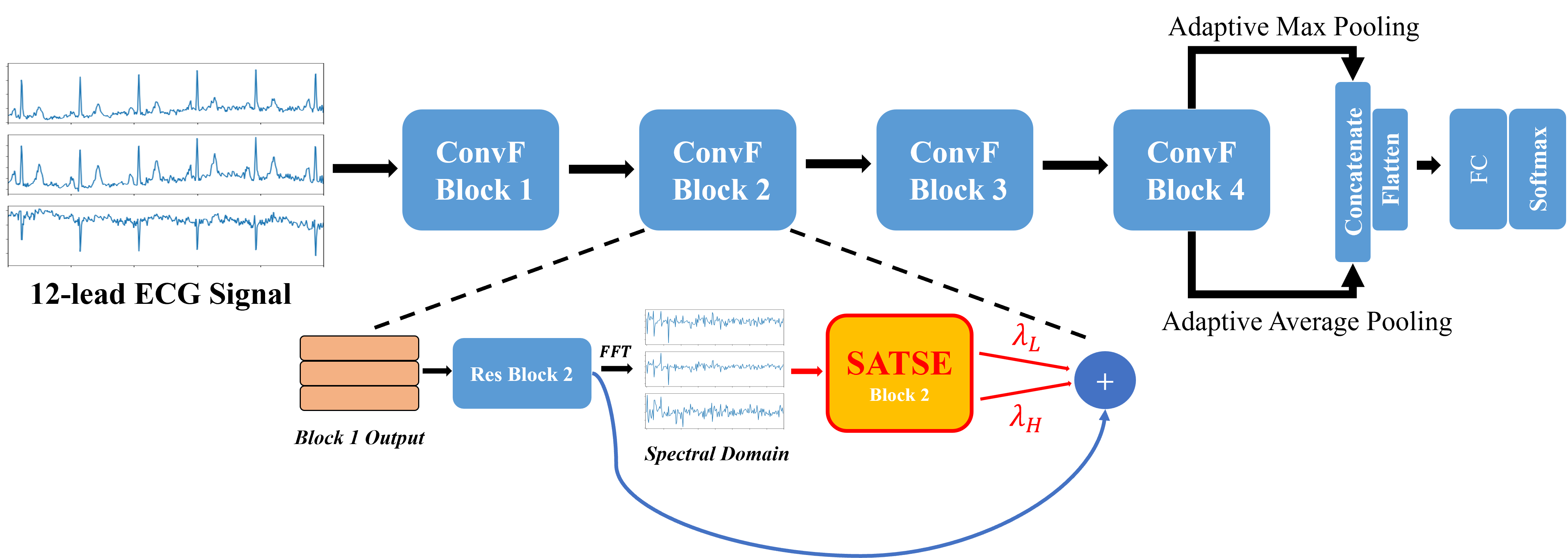}
\caption{Workflow of \ac{SCDNN} where $\lambda_{L}$ and $\lambda_{H}$ denote the coefficients corresponding to low-frequency and high-frequency information}\label{workflow}
\end{figure*}

The rest of this paper is organised as follows. 
Section \ref{method} describes the proposed \ac{SCDNN} with a detailed explanation of information processing in the spectral domain. The numerical results of the proposed network, compared to state-of-the-art approaches, on the $PTB-XL$ and $CPSC2018$ databases are presented in Sec \ref{sec: res of ptbxl} and \ref{sec: res of cpsc}, respectively. We end the paper with a conclusion in Sec \ref{conclusion} where we have also mentioned potential future works.

\section{SCDNN: methodology}
\label{method}

\begin{figure}[htp!]
\centering
\includegraphics[width=0.44\textwidth]{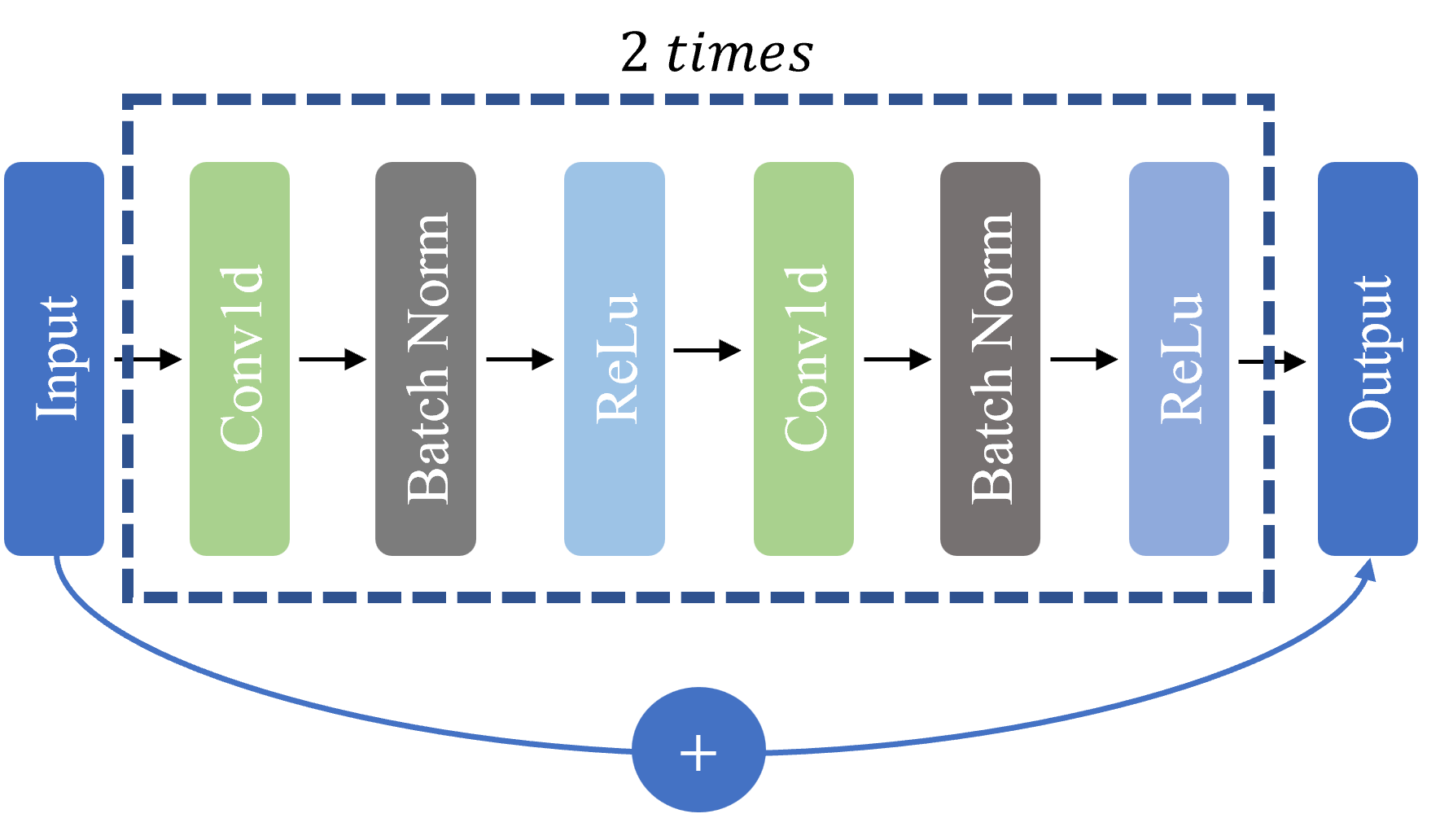}
\caption{Res Block Architecture}\label{resblock}
\end{figure}

In this section, we introduce the workflow of the proposed \ac{SCDNN} with a special focus on the \ac{SATSE} block.
The proposed model is composed of ResNet18~\cite{he2016deep} backbone, four \ac{SATSE} blocks, one adaptive average pooling layer~\cite{liu2016adaptive}, one adaptive max pooling layer~\cite{cho2014finding} and one fully-connected layer as displayed in Fig~\ref{workflow}. The backbone includes $4$ Res blocks (also known as residual blocks~\cite{he2016deep}). The structure of each block is depicted in Fig~\ref{resblock} where all Res blocks used in this study have the same structures with a different number of channels, following the standard ResNet structure~\cite{he2016deep}. Each Res block is combined with one SATSE block to form a \ac{ConvF} component as illustrated in Fig~\ref{resblock}. The \ac{SATSE} block enables the \ac{NN} to learn from the spectral domain.

In each ConvF block, the output of \ac{SATSE} will be added to the  output of the Res block to avoid the information loss from spectral learning~\cite{duhamel1990fast} as illustrated in Fig~\ref{workflow}. Two pooling layers are applied individually on the output of the last ConvF block for dimension reduction. Two pooled features are concatenated with channel dimension, then flatted to a $1D$ vector before passing to the fully-connected layer for classification.

\subsection{Res Block}
    Since untrainable issues have been reported for very deep neural networks~\cite{srivastava2015training}, ResNet was proposed by~\cite{he2016deep} where skip connections between various convolutional layer were adopted to decrease the issue of information vanishing in deep layers of neural networks. A light version of ResNet, namely ResNet18~\cite{he2016identity} is chosen in this work as the backbone. Fig~\ref{resblock} depicts the layout of each Res block, consisting of four $1D$ convolutional layers with a common kernel size and stride as defined in~\cite{he2016deep}. The convolutional layers are employed to extract features across all channels. To enhance the gradient stabilization, a $BatchNorm$ layer~\cite{ioffe2015batch} is deployed after each convolutional layer followed by a $ReLU$~\cite{xu2015empirical} activation function to avoid the gradient vanishing in deep neural networks~\cite{hochreiter2001gradient}.

\subsection{SATSE}
\label{sec:SATSE}
\begin{figure*}[htp!]
\centering
\includegraphics[width=0.8\textwidth]{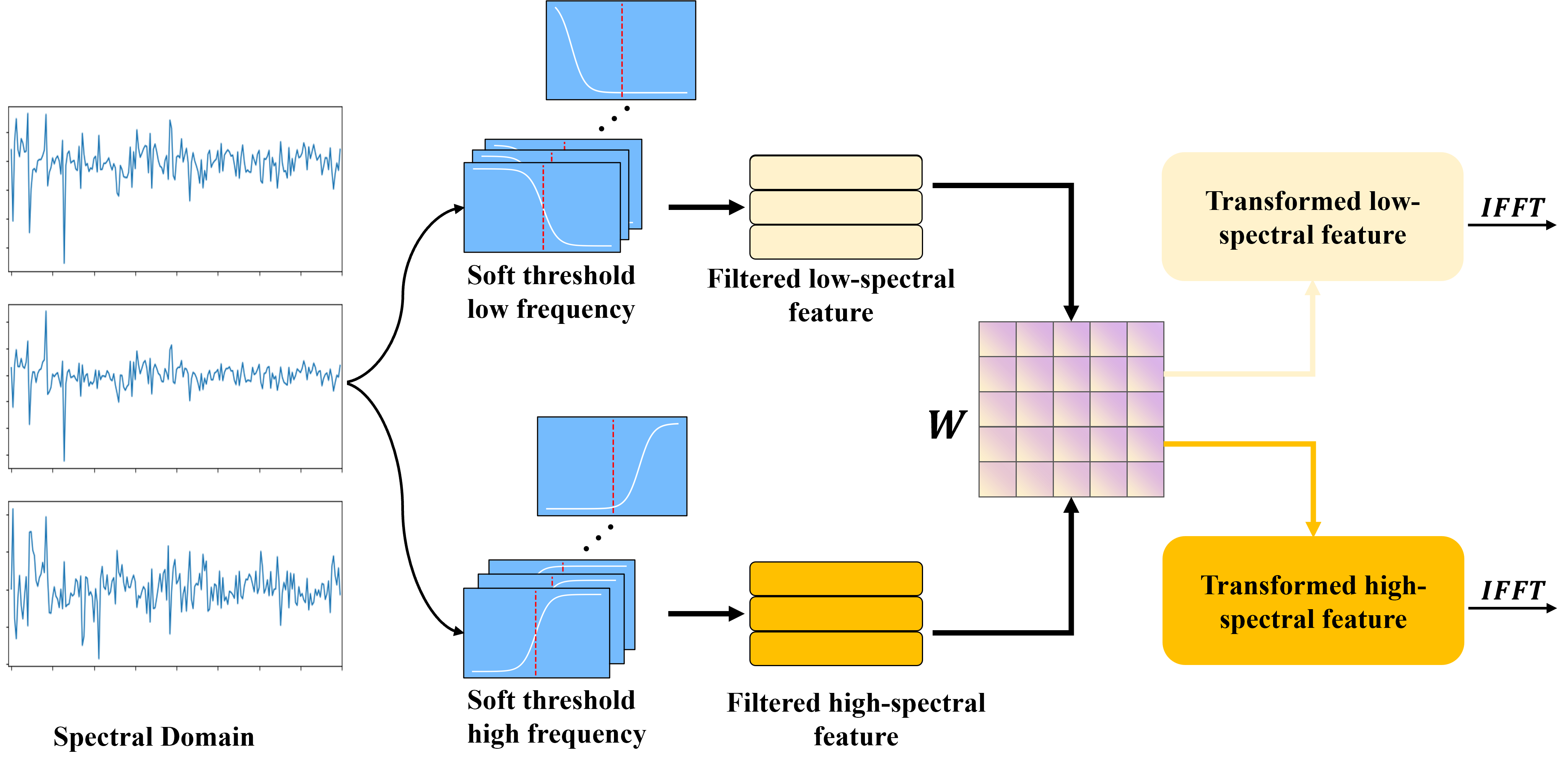}
\caption{SATSE Architecture}\label{SATSE}
\end{figure*}

In this section, we explain in details the soft adaptive threshold and the spectral learning in spectral domain. The pipeline of \ac{SATSE} is illustrated in Fig~\ref{SATSE}. First, \ac{FFT} is deployed in \ac{SATSE} for spectral knowledge converting. The model is thus capable of capturing deep features in time domain and spectral domain simultaneously. To choose the most proper spectral domain, we make use of trainable thresholds in each \ac{SATSE} block to select the spectral modes.
The soft threshold mechanism is capable of avoiding information loss in modes selection since the threshold value is obtained through network back propagation. Additionally, as shown in Fig~\ref{SATSE}, a trainable weight matrix is added to establish connections among different spectral modes. Therefore, those spectral modes in different \ac{SATSE} blocks are determined mutually.   
For the $i^{th} (i=1,...,4)$ \ac{SATSE} block in \ac{SCDNN}, the output of the Res block is denoted as
\begin{align}
    \{f^{(j)}_{i,k}\} \quad \textrm{where} \quad \{j,k \} \in   \{ 0,...,L_i-1\} \times \{ 0,...,C_i-1\} \label{eq:1}
\end{align}
 In Eq \eqref{eq:1}, $C_{i}$ and $ L_{i}$ denote the number of channels and  the signal length of the $i^{th}$ block respectively.
Let $\mathcal{F}$ and $\mathcal{F}^{-1}$ denote the \ac{FFT} and \ac{IFFT} on discrete signals,
the computation of spectral domain features $f^{S,(j)}_{i,k}$ of each \ac{SATSE} block and each channel is performed via discrete Fourier transform,
\begin{align}\label{fourier transform}
    f^{S,(j)}_{i,k} &=  \sum_{n=0}^{L_{i}-1} e^{-\hat{i} \frac{2 \pi}{L_i} n j} f^{(n)}_{i,k}.\\
    \mathcal{F}(f_{i,k} ) &=  [f^{S,(j)}_{i,k}]_{j=0,..,L_i-1}
\end{align}

For the sake of clarity, $\hat{i}$ denotes the imaginary unit in $ e^{-\hat{i} \frac{2 \pi}{L_i}}$ while the index $i$ is referred to the index of the \ac{SATSE} block.
 To compute the soft trainable thresholds, modified sigmoid functions in the high- and low- frequency domain are defined respectively as:
 \begin{align}\label{low threshold}
    \sigma^{L}(x) &= 1 - \frac{1}{1+e^{\gamma_{i}(-x+\varphi_{i} \cdot  L_{i})}}\\
    \sigma^{H}(x) &=  \frac{1}{1+e^{\gamma_{i}(-x+\varphi_{i}\cdot  L_{i})}} \label{high threshold}
\end{align}
 where $\varphi_{i}$ denotes the trainable threshold ratio and $\gamma$ controls the slope of the modified sigmoid function. The dual sigmoid functions $\sigma^{L}$ and $\sigma^{H}$ enable the \ac{SATSE} block to capture knowledge in both low- and high-frequency domains. The use of soft thresholds in Eq \eqref{high threshold} enables the \ac{NN} back-propagation compared to hard threshold functions, i.e.,
  \begin{align}
    \sigma^{L}_\textrm{hard}(x) &= \mathbbm{1}_{x>\varphi_i L_i} \label{hard Lthreshold}\\
    \sigma^{H}_\textrm{hard}(x) &=  \mathbbm{1}_{x\leq\varphi_i L_i} \label{hard Hthreshold}.
\end{align}
 We then obtain filtered high- and low-frequency elements in the spectral domain via
 \begin{align}\label{filtering}
    \widetilde{f^{S^H}_{i,k}} = \sigma^{H}(k) f^{S}_{i,k}, \hspace{5mm} \widetilde{f^{S^L}_{i,k}} = \sigma^{L}(k) f^{S}_{i,k}.
\end{align}
 By definition in Eq \eqref{low threshold} and \eqref{high threshold}, 
 when the values of $\gamma$ (i.e., the slope of the sigmoid functions) are important, $\sigma^{L}(x) \approx \sigma^{L}_\textrm{hard}(x)$ and $\sigma^{H}(x) \approx \sigma^{H}_\textrm{hard}(x)$. Therefore, symmetric results can be obtained when 
 \begin{align}
     \varphi = \lambda \quad \textrm{and} \quad \varphi = 1-\lambda, \quad \forall \lambda \in [0,1]
 \end{align}
  by simply reversing the role of $\sigma^{L}$ and $\sigma^{H}$. Thus the initial value of $\varphi$ is set to be smaller than 0.5 in the training.
 To enable the communication of cross-domain and cross-block information, trainable weight matrices $W_{i,k} \in \mathbb{C}^{4,C_i}$ for different \ac{SATSE} blocks and different channels are employed in the inverse Fourier transformation,
 \begin{align}\label{FC transform}
    \widehat{f^{S^H}_{i,k}} &= W_{i,k} \odot \widetilde{f^{S^H}_{i,k}}, \hspace{5mm} \widehat{f^{S^L}_{i,k}} = W_{i,k} \odot \widetilde{f^{S^L}_{i,k}} \\
    f^{H'}_{i,k} & = \mathcal{F}^{-1}(\widehat{f^{S^H}_{i,k}}) =\frac{1}{L_i} \sum_{n=0}^{L-1} e^{i \frac{2 \pi}{L_i} k n} \widehat{f^{S^H}_{i,n}} \label{IFF1}\\
     f^{L'}_{i,k} & = \mathcal{F}^{-1}(\widehat{f^{S^L}_{i,k}}) =\frac{1}{L_i} \sum_{n=0}^{L-1} e^{i \frac{2 \pi}{L_i} k n} \widehat{f^{S^L}_{i,n}}  \label{IFFT2},
\end{align}
 where $\{ f^{H'}_{i,k}\}$ and $\{ f^{L'}_{i,k}\}$ $ (i = 0...3, k = 0...L_i-1)$ represent the inverse Fourier sequences in time domain. The symbol $\odot$ denotes the element-wise multiplication in Eq \eqref{FC transform}, enabling cross spectral domain communications.
Finally, the low- and high-frequency domain knowledge is converted to time domain through \ac{IFFT} separately, following Eq.~\eqref{IFF1} and~\eqref{IFFT2}. 
We then obtain the output of the \ac{SATSE} block $O_{i}^{\ac{SATSE}}$ by adding the original time domain component $\{f_{i,k}\}$ thanks to two real trainable parameters $\lambda_{L}$ and $\lambda_{H}$,

\begin{align}\label{adjustable coefficient}
    O_{i,k}^{\textrm{SATSE}} &= f_{i,k} + \lambda_{L}f^{L'}_{i,k} + \lambda_{H} f^{H'}_{i,k} \\
        O_{i}^{\textrm{SATSE}} &= [O_{i,k}^\textrm{SATSE}]_{ k \in \{ 0,...,C_i-1\}}. \notag
\end{align}

Instead of a pre-defined threshold (e.g.,~\cite{li2020fourier}),
trainable soft thresholds $\varphi_i$ in the \ac{SATSE} block avoid the information loss due to prior assumptions in \ac{SCDNN}.

The training process of \ac{SCDNN} is summarized in Algorithm \ref{algo-workflow}.

\begin{algorithm}[htp!]
\caption{Training of \ac{SCDNN}}\label{algo-workflow}
\begin{algorithmic}[1]
\Require 12 leads ECG signals and labels \\
Initialize model parameters \\
Set Epochs number $E$, learning rate $\eta$, batch size $m$, number of sequential Res blocks $r$  \\
\For {$k = 0$ to $E$}
    \State Load ECG signals and labels with batch size
    \For {$ i = 0 \hspace{1mm} \textrm{to} \hspace{1mm} r$}
        \State Forward propagation to $\textrm{ResBlock}_{i}$ 
        \State Convert the output from $\textrm{ResBlock}_{i}$ to SD via $FFT$ (cf., Eq \eqref{fourier transform})
        \State Feed SD feature toward $SATSE_{i}$ block
        \State Filtering SD features individually (cf., Eq \eqref{filtering})
        \State Share mutually information on spectral domain (cf., Eq \eqref{FC transform})
        \State Convert features from spectral domain to time domain via $IFFT$ (cf., Eq \eqref{IFF1} and \eqref{IFFT2})
        \State Obtain the output of \ac{SATSE} (cf., Eq \eqref{adjustable coefficient})

    \State Adaptive average and max pooling with Concatenation
    \State Flatten pooled features
    \State Apply softmax to smooth the output probability
    \State Compute the model loss 
    \State optimize model parameters with loss and $\eta$ via Adams optimizer
    \EndFor
\EndFor
\Ensure{Probability of each label}
\end{algorithmic}
\end{algorithm}

\section{Experiments and Analysis}\label{sec: experiment}

\begin{table}[htp!]
\centering
\cl{\caption{Dataset Details of PTB-XL and CPSC2018. This table is taken from~\cite{huang2022snippet}.}
\label{tab: dataset}
\begin{tabular}{|c|c|c|c|}
\hline \hline Datasets & \#Samples & Class & Description \\
\hline & 9528 & NORM & Normal ECG \\
& 5486 & MI & Myocardial Infarction \\
PTB-XL & 5250 & STTC & ST/T Change \\
& 4907 & CD & Conduction Disturbance \\
& 2655 & HYP & Hypertrophy \\
\hline \hline & 918 & Normal & Normal \\
& 1098 & AF & Atrial fibrillation \\
& 704 & I-AVB & First-degree atrioventricular block \\
& 207 & LBBB & Left bundle branch block \\
CPSC2018 & 1695 & RBBB & Right bundle branch block \\
& 556 & PAC & Premature atrial contraction \\
& 672 & PVC & Premature ventricular contraction \\
& 825 & STD & ST-segment depression \\
& 202 & STE & ST-segment elevated \\
\hline \hline
\end{tabular}}
\end{table}

\subsection{Dataset Description}
\label{sec: dataset}
\cl{
In our study, we employ two public 12-leads \ac{ECG} datasets, namely PTB-XL~\cite{wagner2020ptb} and CPSC2018~\cite{liu2018open}, to evaluate the effectiveness of our proposed SCDNN under various scenarios. The details of both datasets are displayed in Table \ref{tab: dataset}.
}
\subsubsection{PTB-XL}
\cl{
The ECG dataset under examination is substantial, encompassing 21,837 ECG signals that were accumulated from 18,885 patients during the period of October 1989 to June 1996. The collected data consists of 12-lead ECGs, each sampled at a rate of 500 Hz with a duration of 10 seconds. Furthermore, each record in this dataset is classified under one of five primary diagnostic categories: Normal (NORM), Myocardial Infarction (MI), ST/T Change (STTC), Conduction Disturbance (CD), and Hypertrophy (HYP).
}
\subsubsection{CPSC2018}
\cl{
This dataset, which is publicly accessible, was accumulated from 11 different hospitals as a part of the 1st China Physiological Signal Challenge. It comprises 6,877 standard 12-lead ECG records, each sampled at a rate of 500 Hz, and the duration of these records ranges from 6 to 60 seconds. The dataset is annotated with nine distinct labels, which include Atrial fibrillation (AF), First-degree atrioventricular block (I-AVB), Left bundle branch block (LBBB), Right bundle branch block (RBBB), Premature atrial contraction (PAC), Premature ventricular contraction (PVC), ST-segment depression (STD), ST-segment elevation (STE), and normal (NORM).
}
\subsection{Implementation}
\label{sec: implement}

\subsubsection{Preprocessing}
\label{sec: preprocess}
\cl{
In every task, we adhere rigorously to the procedure set out in ~\cite{huang2022novel,huang2022snippet}, taking raw ECGs as inputs without any normalization process. The ECGs are resampled to a frequency of 500 Hz during dataset building~\cite{liu2018open,wagner2020ptb}. 
Given the varying signal lengths in the CPSC2018 dataset, we adopt a strategy of padding all ECGs to the maximum length encountered in the dataset, strictly adhering to the procedures outlined in ~\cite{huang2022novel,huang2022snippet}.
}
\subsubsection{Environment}
\cl{
All experiments are implemented on a server with a $12^{\text{th}}$ Intel i7-12700k CPU, 32-GB
memory, and dual NVIDIA GeForce Rtx 3090 GPUs. This server runs an
Ubuntu 20.04 system, and the models are implemented based
on the PyTorch 1.13.1.
}
\subsubsection{Training Parameters Setting}
\label{sec: training setting}
\cl{
For all experimental tasks, we strictly adopt the number of epochs, batch size, initial learning rate, and weight decay rate as mentioned in ~\cite{huang2022snippet,huang2022novel}, set to 50, $32$, $1e^{-4}$, and $2e^{-5}$ respectively. To establish an fair comparison with baseline methods, we select the standard $CrossEntropyLoss$~\cite{de2005tutorial} and Adams~\cite{snoek2012practical} as the loss function and the optimizer for the training process. In adherence to the approach outlined in ~\cite{huang2022snippet,huang2022novel}, we perform a ten-fold reduction of the learning rate at the 20th epoch. Within the \ac{SATSE} blocks, the initial settings of $\varphi_{i}$ and $\gamma_{i}$ are established at $0.4$ and $0.5$ per block, whereas $\lambda_L$ and $\lambda_H$ are initialized at $0$. These parameters are subsequently updated during the network back-propagation.
}

\subsection{Metrics of Performance}
\label{sec: metric}
\cl{
In order to evaluate the efficiency of the SCDNN for ECG classification, various performance measures are taken into account. While the F1 score is one of the key metrics, precision, recall, and accuracy are also critical for validating the performance of the models.
All the aforementioned metrics, F1-score, Precision, Recall, and Accuracy are calculated using macro averaging.
}
\begin{enumerate}
    \item \cl{F1-score: This is the harmonic mean of precision and recall, offering a balanced measure especially when classes are unevenly distributed.}

    \item \cl{Precision: This measures the percentage of true positive predictions among all positive predictions, indicating the level of false-positive error.}

    \item \cl{Recall: Also known as sensitivity, it shows the proportion of actual positives correctly identified, reflecting the model's ability to detect all positive cases.}

    \item \cl{Accuracy: This is the ratio of correct predictions (both positive and negative) to the total number of instances, indicating the overall correctness of the model.}
\end{enumerate}

\subsection{Results on PTB-XL}
\label{sec: res of ptbxl}
\cl{An in-depth analysis of the classification results on the PTB-XL dataset, as presented in Table \ref{tab: ptb res}, demonstrates the superiority of SCDNN over the competing methods across all metrics, including accuracy, precision, recall, and F1-score. This analysis takes into account both the mean and the standard deviation of the results, adding depth and precision to the assessment.}

\cl{SCDNN emerges as the top performer in terms of accuracy, boasting a high score of 0.835 and a notably low standard deviation of 0.003. This significantly outperforms the closest competitor, SPN-V2, which achieves an accuracy of 0.805, but with a higher standard deviation of 0.021, thus highlighting the consistent accuracy of the SCDNN model.}

\cl{In terms of precision, SCDNN outshines the competition once again, achieving a score of 0.804, accompanied by an impressively low standard deviation of 0.005. The next best method, SPN-V2, achieves a slightly lower precision of 0.79, but with a higher standard deviation of 0.020.}

\cl{The recall metric also underlines SCDNN's superiority, with a high score of 0.787 and a small standard deviation of 0.006. CKNA, while achieving a respectable recall of 0.756, does so with a considerably higher standard deviation of 0.03.}

\cl{Regarding the F1-score, which offers a balanced measure of both precision and recall, SCDNN again leads the pack, achieving a score of 0.792 and maintaining a low standard deviation of 0.005 in Table \ref{tab: ptb res}. The closest competitors, CKNA and SPN-V2, both present a lower F1-score of 0.762, but with higher standard deviations (0.026 and 0.027 respectively).}

\cl{In conclusion, the consistent high performance of SCDNN in all the metrics considered on Table \ref{tab: ptb res} underscores its robustness and effectiveness in ECG signal classification. Furthermore, SCDNN's consistently low standard deviation scores attest to its stability and reliability as a method, making it a promising candidate for further exploration and application in the field.
}

\begin{table*}[htp!]
\centering
\cl{
\caption{Classification Results on PTB-XL. All results are inherited from~\cite{huang2022novel,huang2022snippet}.}
\label{tab: ptb res}
\begin{tabular}{|c|c|c|c|c|}
\hline \hline & Accuracy  & Precision & Recall & F1-score  \\
\hline \hline SR2-CF2~\cite{SR2-CF2} & $0.229 \pm 0.003$  & $0.227 \pm 0.140$ & $0.200 \pm 0.001$ & $0.076 \pm 0.001$  \\
\hline EARLIEST~\cite{EARLIEST} & $0.454 \pm 0.009$ & $0.351 \pm 0.056$ & $0.227 \pm 0.007$ & $0.188 \pm 0.010$  \\
\hline TEASER~\cite{TEASER} & $0.584 \pm 0.007$  & $0.494 \pm 0.009$ & $0.455 \pm 0.009$ & $0.461 \pm 0.009$  \\
\hline MDDNN~\cite{MDDNN} & $0.733 \pm 0.017$  & $0.622 \pm 0.041$ & $0.581 \pm 0.025$ & $0.586 \pm 0.027$  \\
\hline ETEeTSC~\cite{ETEeTSC} & $0.744 \pm 0.015$  & $0.653 \pm 0.025$ & $0.620 \pm 0.029$ & $0.629 \pm 0.027$  \\
\hline SPN~\cite{SPN} & $0.794 \pm 0.013$  & $0.732 \pm 0.025$ & $0.705 \pm 0.031$ & $0.710 \pm 0.023$  \\
\hline SPN-V2~\cite{huang2022snippet} & $ 0.805 \pm 0.021$  & $0.790 \pm 0.020$ & $0.744 \pm 0.031$ & $0.762 \pm 0.027$  \\
\hline CKNA~\cite{huang2022novel} & $0.799 \pm 0.022$ & - & $0.756 \pm 0.030$ & $0.762 \pm 0.026$  \\
\hline \hline SCDNN & $\textbf{0.835} \pm \textbf{0.003}$ & $\textbf{0.804} \pm \textbf{0.005}$ & $\textbf{0.787} \pm \textbf{0.006}$ & $\textbf{0.792} \pm \textbf{0.005}$   \\
\hline \hline
\end{tabular}
}
\end{table*}

\subsection{Results on CPSC2018.}
\label{sec: res of cpsc}
\cl{Analysing the results from the CPSC2018 dataset from Table \ref{tab: cpsc res}, it is apparent that the SCDNN method significantly outclasses the other methodologies in all examined metrics, mirroring its superior performance in the PTB-XL dataset. This consistent superior performance across disparate datasets underscores the high degree of generalizability inherent in the SCDNN method, confirming its adaptability to diverse data sources and categories.}

\cl{In relation to accuracy, SCDNN achieves the highest score of 0.859, accompanied by a remarkably low standard deviation of 0.003. The nearest rival, the CKNA method, achieves an accuracy of 0.812 but with a significantly higher standard deviation of 0.013. This suggests SCDNN's superior and consistent predictive capability across varying datasets.}

\cl{The trend continues in the precision metric, where SCDNN attains a superior score of 0.838, again exhibiting a low standard deviation of 0.012. The nearest competitor, the SPN method, achieves a precision of 0.762, a substantial difference that underlines the prowess of SCDNN in accurately classifying positive instances.}

\cl{Furthermore, the SCDNN method surpasses all other methods in terms of recall, with an outstanding score of 0.781 and an extremely low standard deviation of 0.003. This significantly surpasses the next best recall score of 0.764 achieved by the CKNA method. This evidence suggests that SCDNN excels in correctly identifying a higher proportion of actual positive instances.}

\cl{Finally, regarding the F1-score, which reflects the balance between precision and recall, SCDNN continues to outperform with a score of 0.782 and a low standard deviation of 0.003. In comparison, the second-highest F1-score is 0.764, held by the CKNA method.}

\cl{In summary, the superior performance of the SCDNN method across all evaluated metrics in the CPSC2018 dataset, mirroring its performance on the PTB-XL dataset, highlights its powerful generalizability and transferability. This suggests the SCDNN method's considerable promise in the realm of ECG signal analysis.
}

\begin{table*}[htp!]
\centering
\cl{
\caption{Results on CPSC2018. All results are inherited from~\cite{huang2022novel,huang2022snippet}.}
\label{tab: cpsc res}
\begin{tabular}{|c|c|c|c|c|}
\hline \hline & Accuracy  & Precision & Recall & F1-score  \\
\hline \hline SR2-CF2~\cite{SR2-CF2} & $0.167 \pm 0.009$ & $0.579 \pm 0.110$ & $0.160 \pm 0.010$ & $0.109 \pm 0.016$  \\
\hline EARLIEST~\cite{EARLIEST} & $0.282 \pm 0.012$  & $0.168 \pm 0.055$ & $0.150 \pm 0.015$ & $0.114 \pm 0.018$  \\
\hline TEASER~\cite{TEASER} & $0.584 \pm 0.018$ & $0.569 \pm 0.057$ & $0.465 \pm 0.017$ & $0.480 \pm 0.020$  \\
\hline MDDNN~\cite{MDDNN} & $0.585 \pm 0.015$ & $0.522 \pm 0.016$ & $0.511 \pm 0.015$ & $0.511 \pm 0.015$  \\
\hline ETEeTSC~\cite{ETEeTSC} & $0.733 \pm 0.021$ & $0.695 \pm 0.030$ & $0.668 \pm 0.022$ & $0.672 \pm 0.026$ \\
\hline SPN~\cite{SPN} & $0.788 \pm 0.015$ & $0.762 \pm 0.018$ & $0.742 \pm 0.014$ & $0.745 \pm 0.015$ \\
\hline SPN-V2~\cite{huang2022snippet} & $0.786 \pm 0.017$ & $0.741 \pm 0.027$ & $0.724 \pm 0.026$ & $0.727 \pm 0.026$ \\
\hline CKNA~\cite{huang2022novel} & $0.812 \pm 0.013 $ & - & $0.764 \pm 0.026$ & $0.764 \pm 0.023$ \\
\hline \hline SCDNN & $\textbf{0.859} \pm \textbf{0.003}$ & $\textbf{0.838} \pm \textbf{0.012}$ & $\textbf{0.781} \pm \textbf{0.003}$ & $\textbf{0.782} \pm \textbf{0.003}$ \\
\hline \hline
\end{tabular}
}
\end{table*}

\subsection{Model Efficiency Analysis}
\begin{figure*}[htp!]
\centering
\subfloat[\label{ptb_infer}Inference Time on and F1-score PTB-XL.]{\includegraphics[width = 0.45\textwidth]{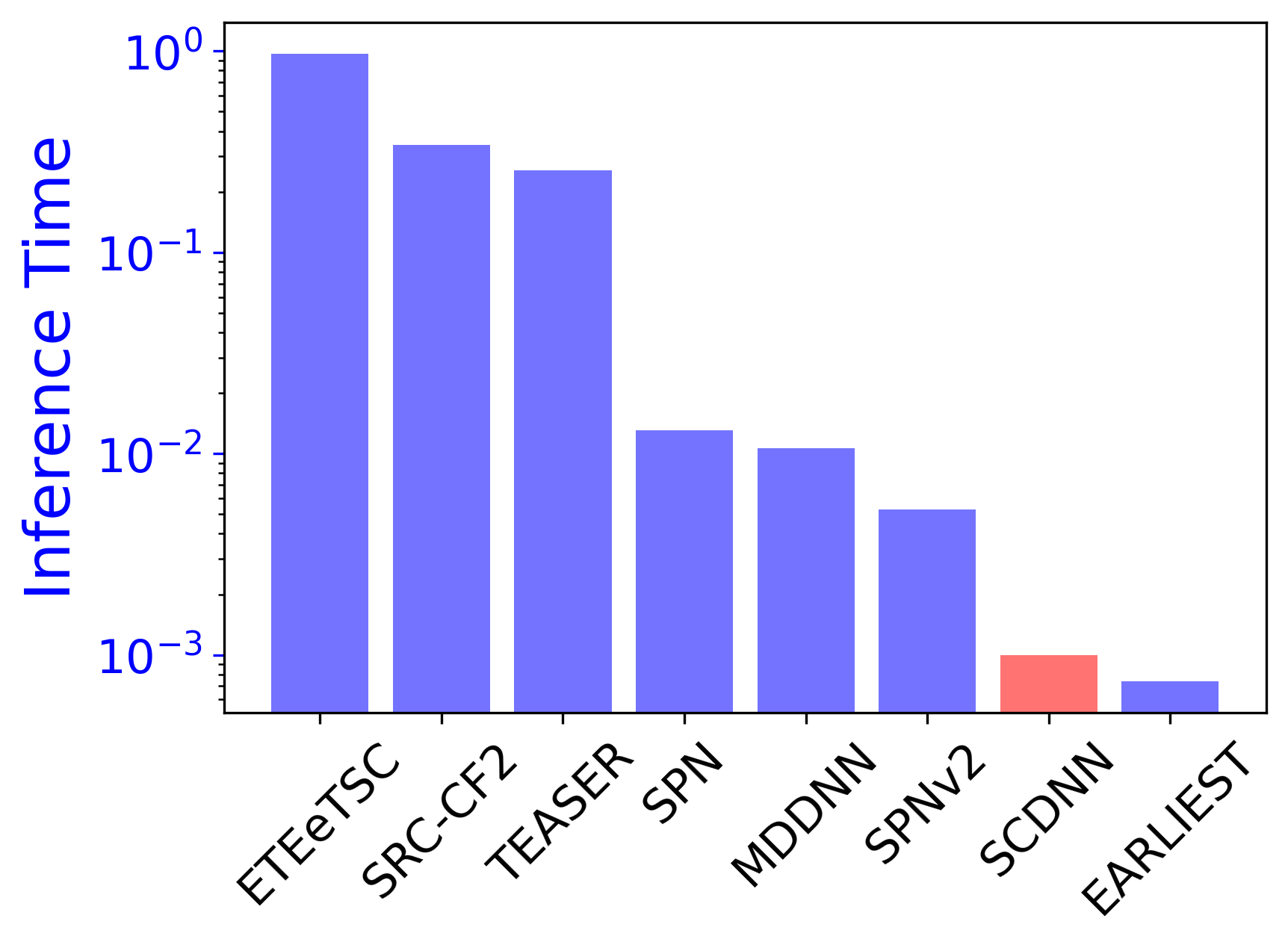}}
\hspace{7mm}
\subfloat[\label{ptb_acc}Accuracy of all methods on PTB-XL.]{\includegraphics[width = 0.5\textwidth]{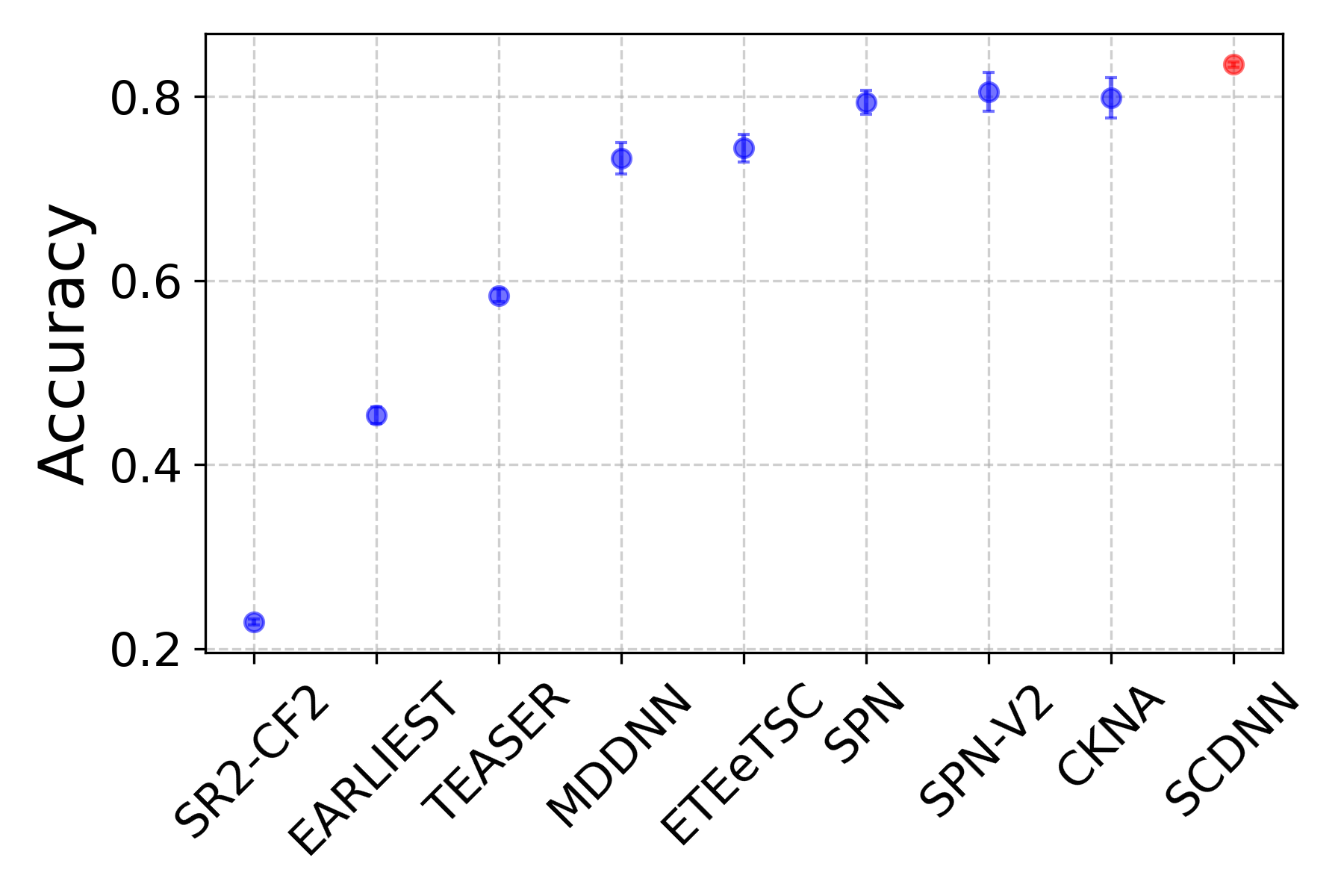}}\\
\subfloat[\label{ptb_f1}F1-score of all methods on PTB-XL.]{\includegraphics[width = 0.5\textwidth]{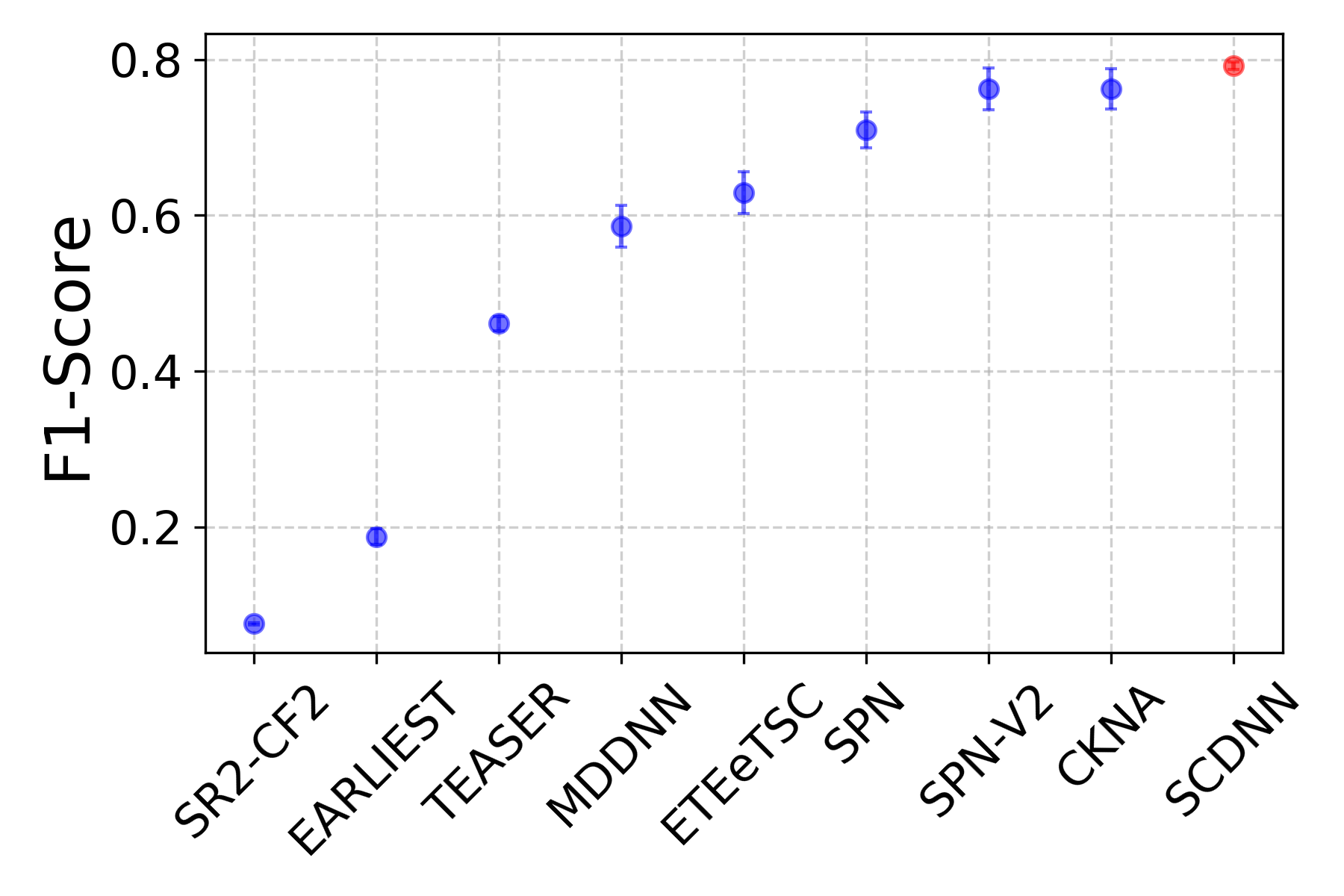}}
\subfloat[\label{ptb_rec}Recall of all methods on PTB-XL.]{\includegraphics[width = 0.5\textwidth]{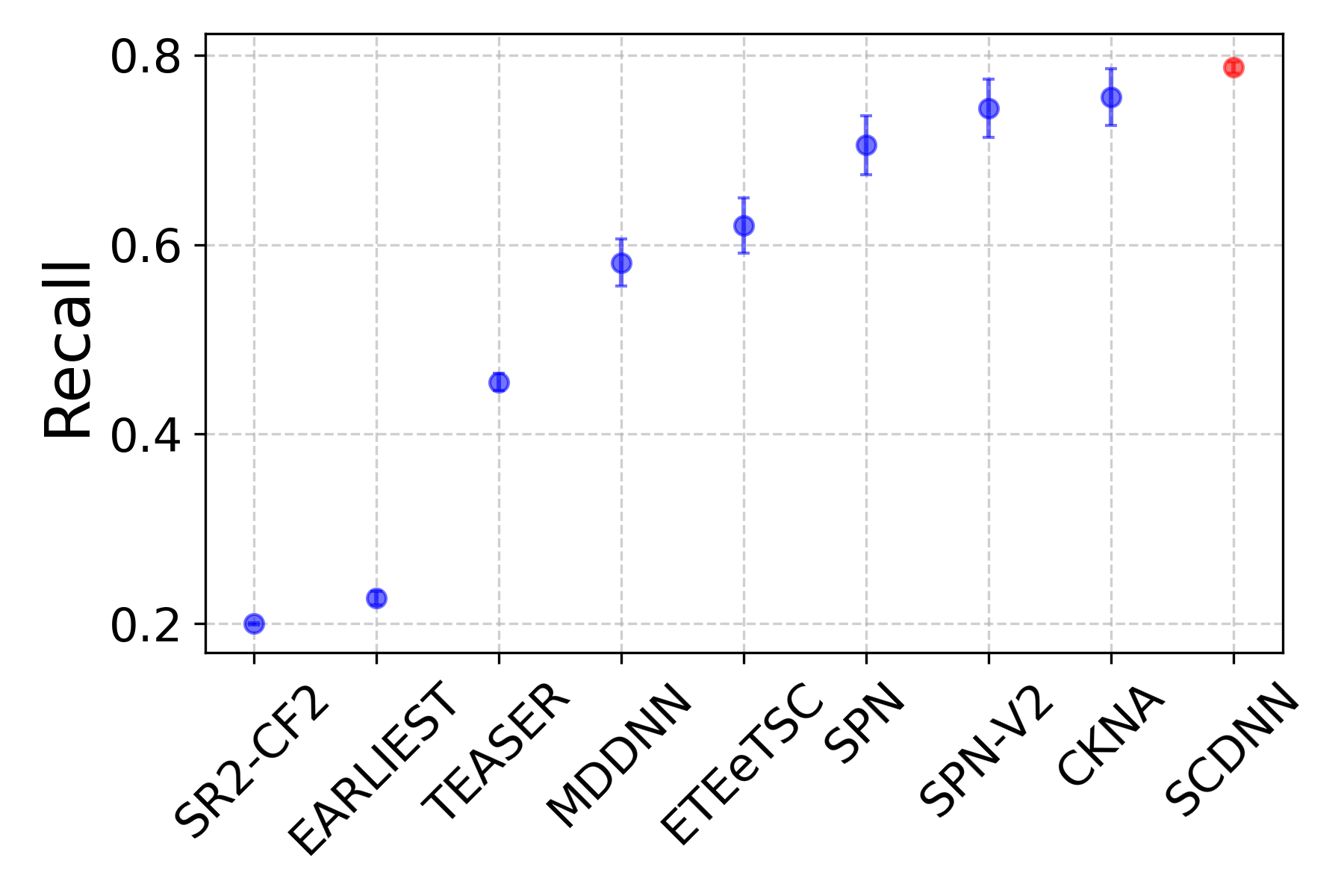}}
 
   \caption{Performance regarding inference time and prediction performance of SCDNN compared to baseline approaches on the PTB-XL dataset.}
   \label{fig: infer metric ptb}
\end{figure*}

\begin{figure*}[htp!]
\centering
\subfloat[\label{cpsc_infer}Inference Time and F1-score on CPSC2018.]{\includegraphics[width = 0.42\textwidth]{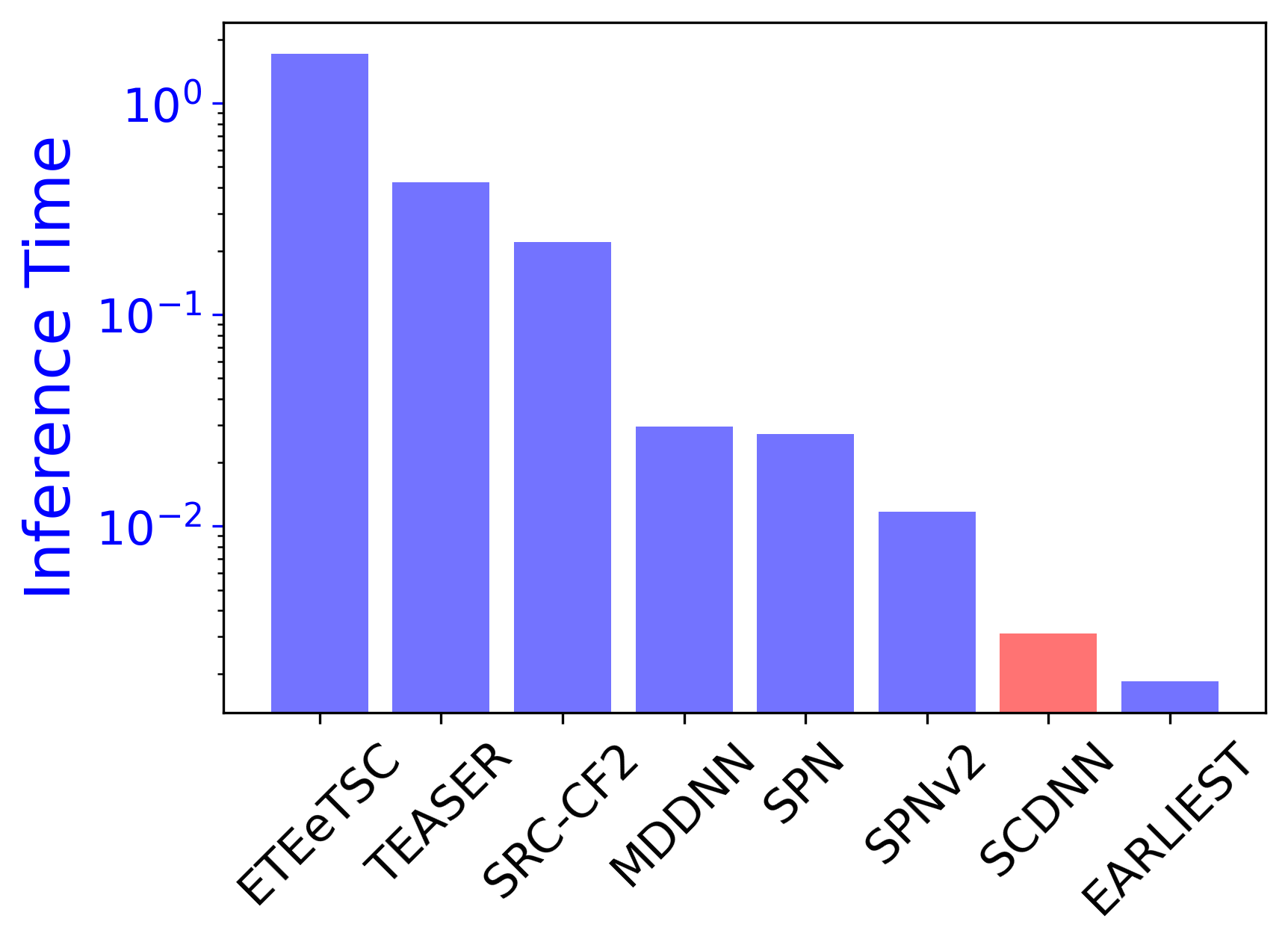}}
\subfloat[\label{cpsc_acc}Accuracy of all methods on CPSC2018.]{\includegraphics[width = 0.47\textwidth]{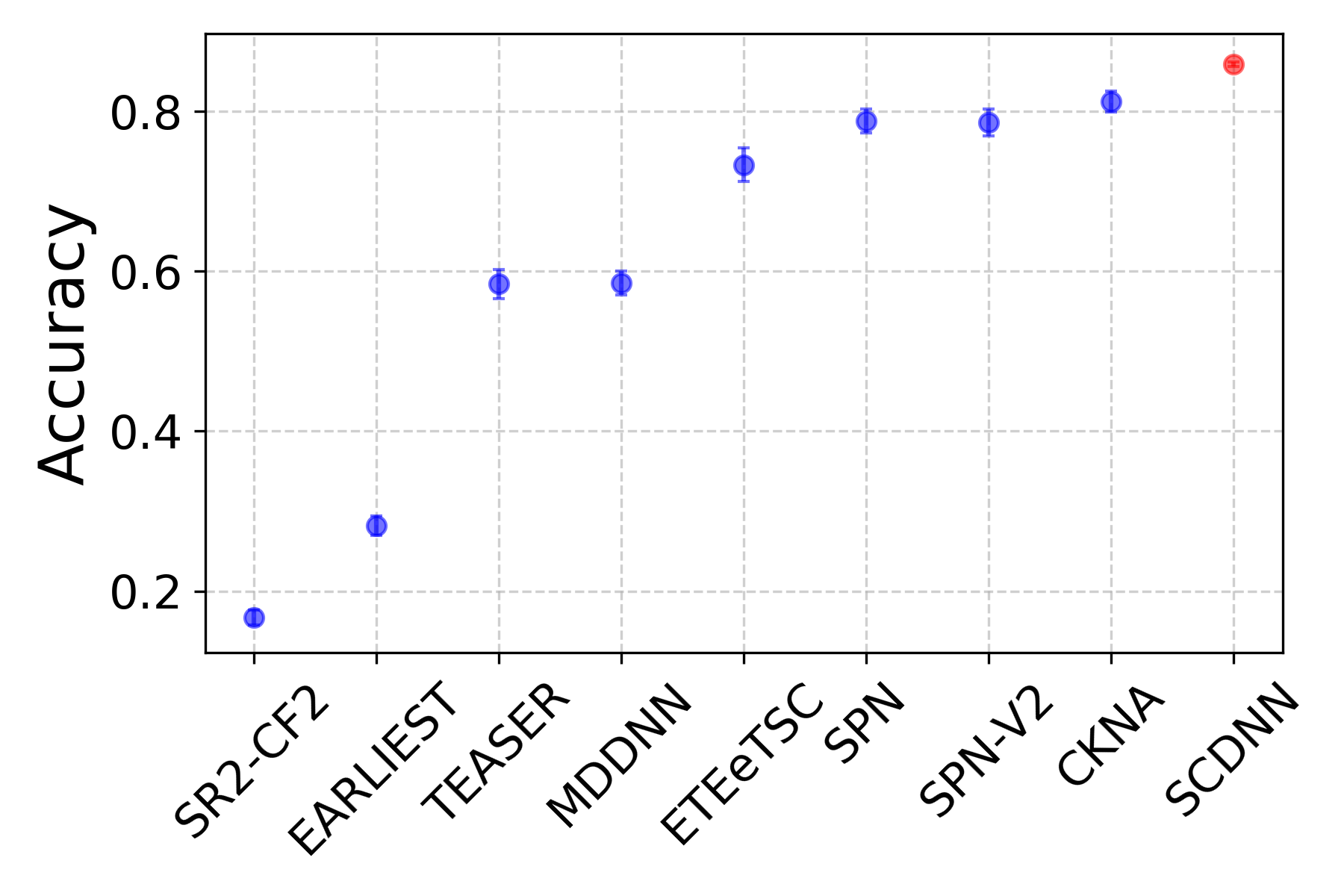}}\\
\subfloat[\label{cpsc_f1}F1-score of all methods on CPSC2018.]{\includegraphics[width = 0.44\textwidth]{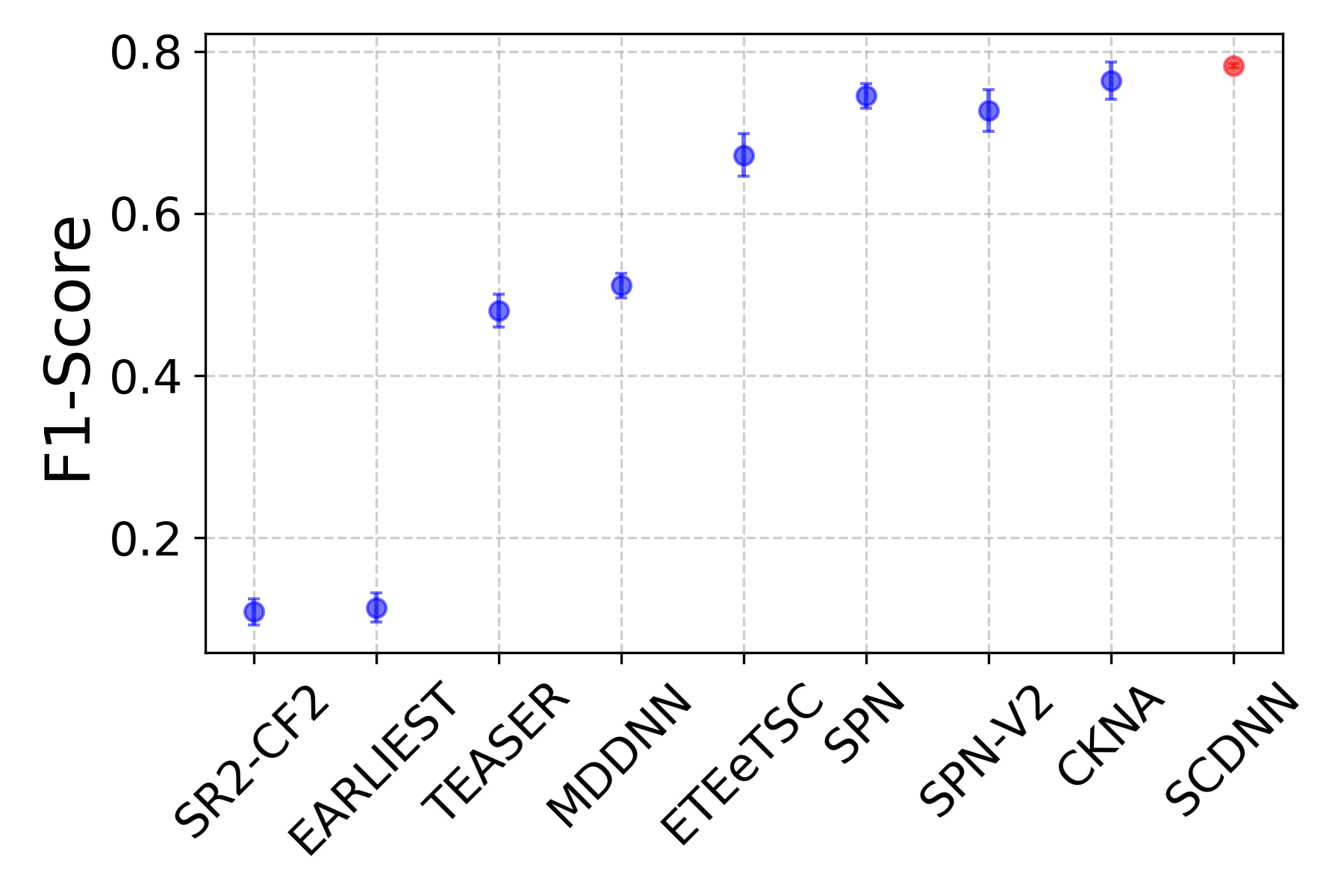}}
\subfloat[\label{cpsc_rec}Recall of all methods on CPSC2018.]{\includegraphics[width = 0.44\textwidth]{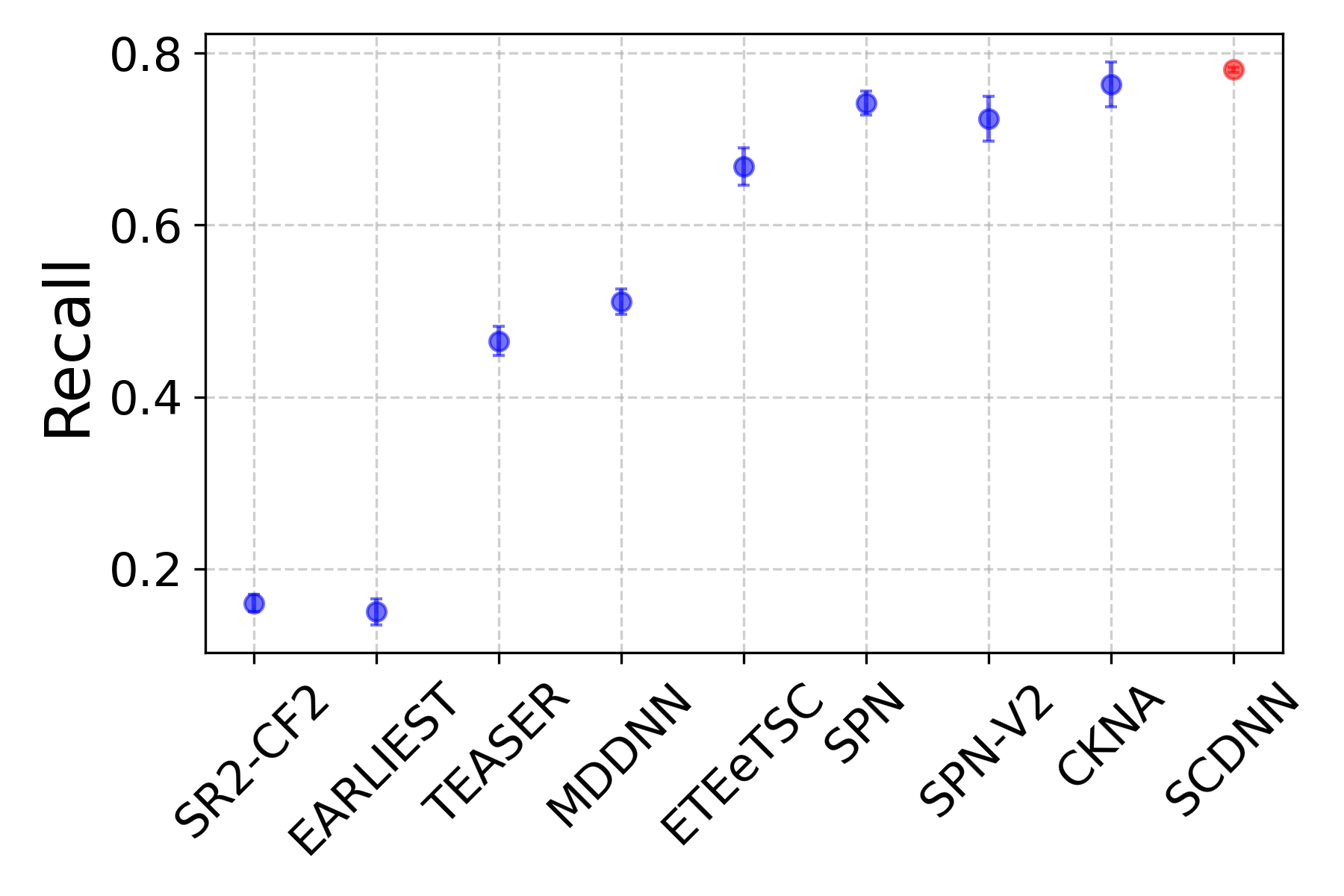}}
 
   \caption{Performance regarding inference time and prediction performance of SCDNN compared to baseline approaches on the CPSC2018 dataset.}
   \label{fig: infer metric cpsc}
\end{figure*}

\cl{For a comparative analysis of SCDNN's efficiency comparing to other benchmarks, we present Fig \ref{ptb_infer} and \ref{cpsc_infer}, charting the inference time on PTB-XL and CPSC2018 datasets respectively. Further, performance metric, including accuracy, recall, and f1-score, are delineated in Fig \ref{ptb_acc}-\ref{ptb_rec} and \ref{cpsc_acc}-\ref{cpsc_rec}.
An inspection of Fig \ref{ptb_infer} and \ref{cpsc_infer} reveals that the SCDNN model exhibits the second lowest inference duration among all compared models, thereby conferring a significant temporal efficiency advantage in the inference phase when juxtaposed against other baselines. Meanwhile, Fig \ref{ptb_acc}-\ref{ptb_rec} and \ref{cpsc_acc}-\ref{cpsc_rec} elucidate the fact that SCDNN consistently achieves the highest performance across all three metrics, across both datasets, a testament to its superior proficiency.}

\cl{Interestingly, the model that attains the highest inference speed, EARLIEST, only delivers second lowest results, highlighting the noteworthy trade-off between computational speed and performance. Nevertheless, SCDNN effectively bridges this gap, striking an optimal balance between temporal efficiency and performance metrics. As such, the empirical evidence underscores the unrivalled efficacy of SCDNN, making it a commendable choice in both time efficiency and performance arenas.
}

\begin{figure*}[htp!]
\centering
\subfloat[\label{ptb_radar}Recall of 5 diseases on PTB-XL.]{\includegraphics[width = 0.47\textwidth]{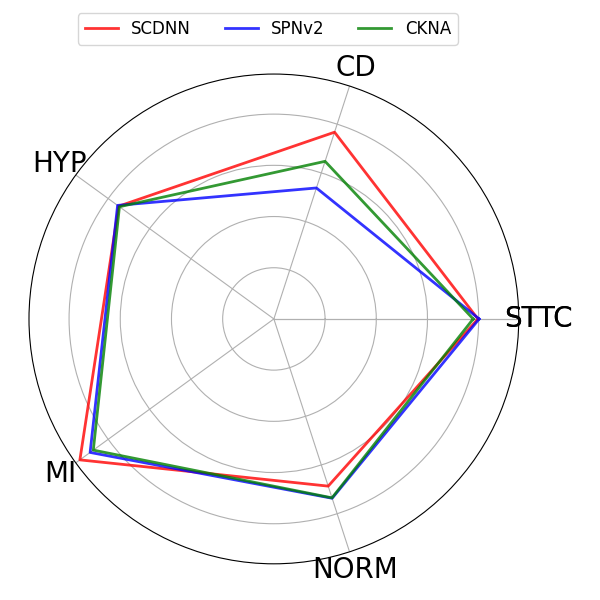}}
\subfloat[\label{cpsc_radar}Recall of 9 diseases on CPSC2018.]{\includegraphics[width = 0.47\textwidth]{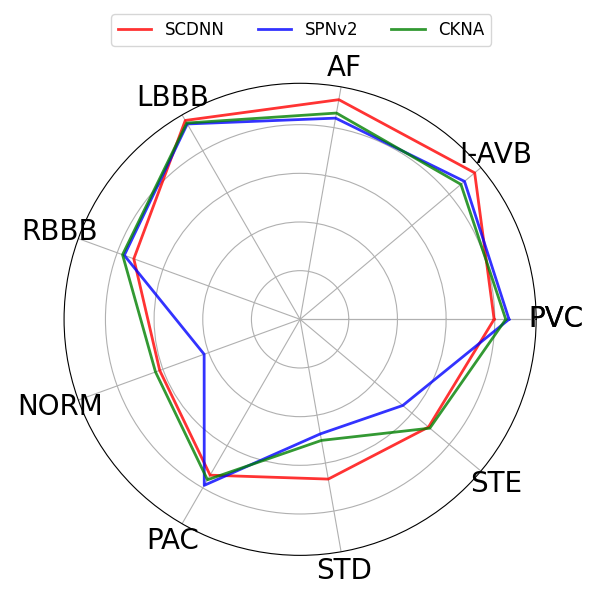}}
 
   \caption{Recall on each disease of SCDNN compared to the best two baseline.}
   \label{fig: radar}
\end{figure*}

\cl{
In a bid to further compare the performance characteristics of SCDNN, SPNv2, and CKNA--the two best performing baselines--we have charted the recall of each disease on both the PTB-XL and CPSC2018 datasets, as depicted in Figure \ref{fig: radar}. In the context of the PTB-XL dataset, SCDNN establishes superiority over other methodologies on CD and MI, while attaining results commensurate with other diseases. When considering the CPSC2018 dataset, SCDNN surpasses other baseline methodologies on AF, I-AVB, STD, with outcomes on remaining diseases being in the same range as other methods. These findings highlight the robust and consistent performance of the SCDNN model across various conditions and datasets, underscoring its potential as an effective tool for cardiovascular disease detection and classification.
}

\subsection{Impact of Adaptive Frequency Filter}
\begin{figure*}[htp!]
\centering

 \includegraphics[width = 0.95\textwidth]{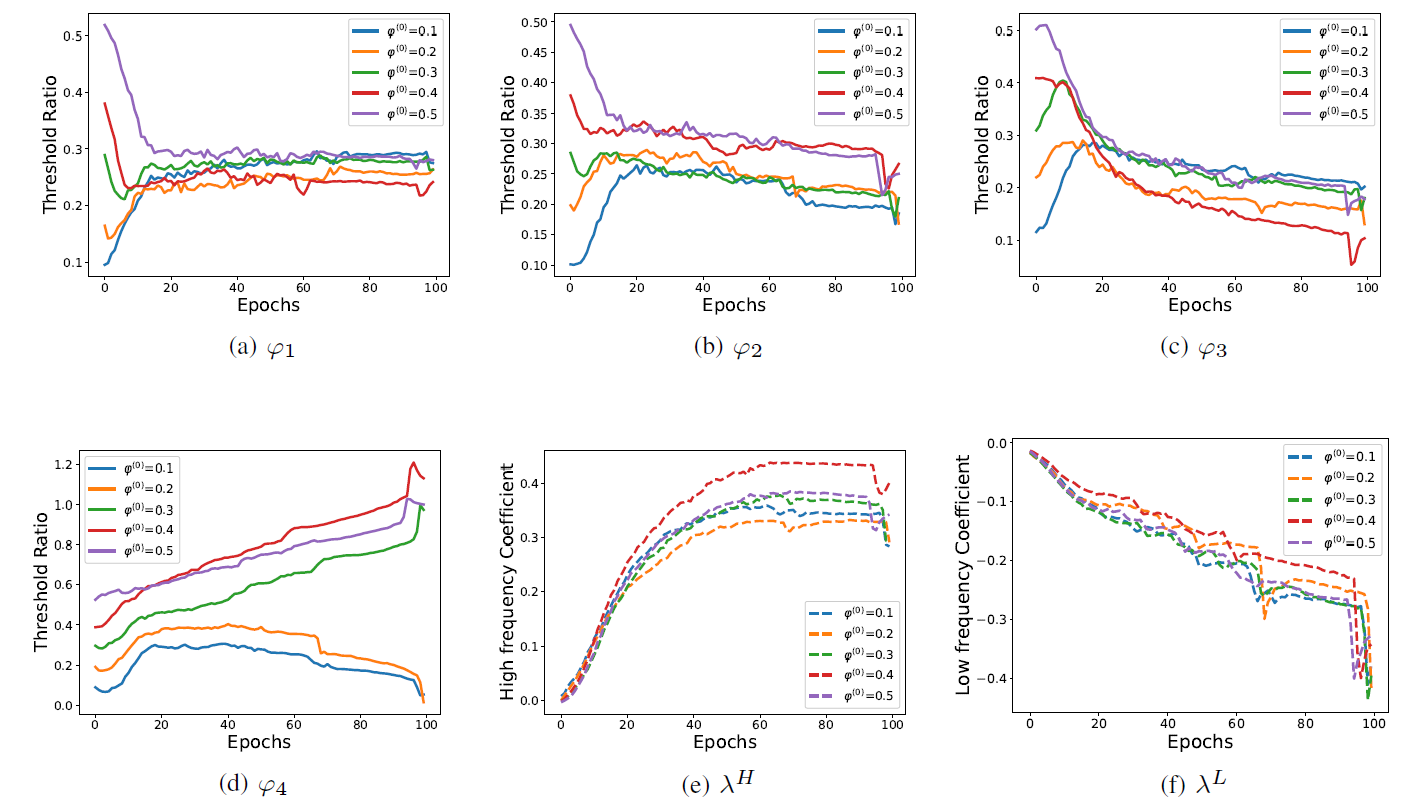}
   \caption{Evolution of $\varphi_i, i =1,..,4, \lambda^H$ and $\lambda^L$ against training Epochs with different initial values of $\varphi_i$ on PTB-XL dataset.}
   \label{fig:param}
\end{figure*}

\cl{The evolution of model parameters against training Epochs is illustrated in Fig~\ref{fig:param}. 
Regardless the initial values, the convergence can be observed  
for $\varphi_{1},\varphi_{2}$ and $\varphi_{3}$ in Fig~\ref{fig:param}(a-c). On the other hand, $\varphi_{4}$ converges numerically to 0 or 1 regarding different initial values as shown in Fig~\ref{fig:param} (d). In fact, as explained in Section \ref{sec:SATSE}, $\varphi_{i} \rightarrow 0$ and $\varphi_{i} \rightarrow 1$, this will lead to the same numerical outputs for the neural network by simply reversing the role of $\sigma^{L}$ and $\sigma^{H}$. We also display the evolution of $\lambda^H$ and $\lambda^L$ accordingly in Fig~\ref{fig:param}(e,f). Regardless the initial values of $\varphi^i$, the evolution of $\lambda^H$ and $\lambda^L$ remains numerically stable. These results underline the robustness of the proposed \ac{SCDNN} with a large range of initial parameters. 
}

\subsection{Ablation Study}
\subsubsection{Impact of SATSE Blocks}
\cl{In the ablation study presented in Table \ref{tab:ablation_study_block}, we examine the impact of varying numbers of SATSE blocks on model performance across two datasets, PTB-XL and CPSC2018.
In the absence of any SATSE blocks, the model yields the lowest performance. The addition of a single SATSE block significantly enhances the model's effectiveness. The upward trend continues with the integration of two and three SATSE blocks, showing consistent improvements in all metrics.
Notably, the model incorporating all four SATSE blocks achieves the highest performance on both datasets. This outcome underlines the cumulative effectiveness of SATSE blocks, indicating a directly proportional relationship between the number of SATSE blocks and model performance. Thus, SATSE blocks have been validated as a crucial component for improving model performance on both the PTB-XL and CPSC2018 datasets.}

\begin{table*}[htp!]
    \caption{
    Ablation study of the number of SATSE blocks.
    }
    \centering
    \cl{
    \scalebox{0.85}{
    \begin{tabular}{c c c c c|  c c c c|  c c c c}
    \hline \hline
    \multicolumn{5}{c|}{SATSE Blocks} & \multicolumn{4}{c|}{PTB-XL} &  \multicolumn{4}{c}{CPSC2018}\\
    None & 1 & 2 & 3 & 4 & Accuracy & Precision & Recall & F1-score & Accuracy & Precision & Recall & F1-score\\
    \hline
      \checkmark& & & & & $0.711 \pm 0.012$ & $0.677 \pm 0.011$ & $0.668 \pm 0.014$ & $0.673 \pm 0.013$ & $0.727 \pm 0.016$ & $0.708 \pm 0.015$ & $0.697 \pm 0.018$ & $0.702 \pm 0.017$\\
      & \checkmark & & & & $0.754 \pm 0.006$ & $0.725 \pm 0.007$ & $0.714 \pm 0.008$ & $0.719 \pm 0.008$ & $0.776 \pm 0.007$ & $0.755 \pm 0.010$ & $0.746 \pm 0.011$ & $0.750 \pm 0.011$\\
      & \checkmark & \checkmark & & & $0.786 \pm 0.018$ & $0.759 \pm 0.017$ & $0.744 \pm 0.020$ & $0.751 \pm 0.019$ & $0.811 \pm 0.021$ & $0.790 \pm 0.024$ & $0.777 \pm 0.025$ & $0.783 \pm 0.024$\\
      & \checkmark & \checkmark & \checkmark & & $0.814 \pm 0.004$ & $0.786 \pm 0.005$ & $0.772 \pm 0.006$ & $0.779 \pm 0.005$ & $0.837 \pm 0.005$ & $0.819 \pm 0.008$ & $0.806 \pm 0.007$ & $0.812 \pm 0.007$\\
      & \checkmark & \checkmark & \checkmark & \checkmark & $\textbf{0.835} \pm \textbf{0.003}$ & $\textbf{0.804} \pm \textbf{0.005}$ & $\textbf{0.787} \pm \textbf{0.006}$ & $\textbf{0.792} \pm \textbf{0.005}$ & $\textbf{0.859} \pm \textbf{0.003}$ & $\textbf{0.838} \pm \textbf{0.012}$ & $\textbf{0.781} \pm \textbf{0.003}$ & $\textbf{0.782} \pm \textbf{0.003}$\\
    \hline \hline
    \end{tabular}
    }}
    \label{tab:ablation_study_block}
\end{table*}

\subsubsection{Impact of Fixed Frequency Threshold}
\cl{
In our investigation, replacing adaptive $\varphi_{i}$ with fixed values ranging from 0.1 to 0.4, we observe notable performance variations on the PTB-XL and CPSC2018 datasets as demonstrated in Table \ref{tab:ablation_study_phi}.
The ablation study in Table \ref{tab:ablation_study_phi} shows that a fixed $\varphi_{i}$ of 0.2 delivers the highest performance across all metrics. However, performance diminishes slightly as $\varphi_{i}$ increases beyond this point, which underscores the sensitivity of model outcomes to the selected $\varphi_{i}$ value.
When comparing this to Tables \ref{tab: cpsc res} and \ref{tab: ptb res}, we find that our SCDNN model with an adaptive $\varphi_{i}$ outperforms the model using a fixed $\varphi_{i}$ of 0.2. This superior performance of the adaptive $\varphi_{i}$ demonstrates its capability to effectively adapt to different data variations, reinforcing the advantage of flexibility in parameter configuration.
}

\begin{table*}[htp!]
    \caption{
    Ablation study of the fixed $\varphi_{i}$ value.
    }
    \centering
    \cl{
    \scalebox{0.85}{
    \begin{tabular}{c c c c|  c c c c|  c c c c}
    \hline \hline
    \multicolumn{4}{c|}{Fixed $\varphi$} & \multicolumn{4}{c|}{PTB-XL} &  \multicolumn{4}{c}{CPSC2018}\\
    0.1 & 0.2 & 0.3 & 0.4 & Accuracy & Precision & Recall & F1-score & Accuracy & Precision & Recall & F1-score\\
    \hline
      \checkmark & & & & $0.792 \pm 0.007$ & $0.765 \pm 0.008$ & $0.750 \pm 0.010$ & $0.757 \pm 0.009$ & $0.815 \pm 0.008$ & $0.796 \pm 0.011$ & $0.781 \pm 0.012$ & $0.789 \pm 0.011$\\
      & \checkmark & & & $\textbf{0.822} \pm \textbf{0.006}$ & $\textbf{0.791} \pm \textbf{0.007}$ & $\textbf{0.774} \pm \textbf{0.008}$ & $\textbf{0.781} \pm \textbf{0.006}$ & $\textbf{0.843} \pm \textbf{0.006}$ & $\textbf{0.824} \pm \textbf{0.011}$ & $\textbf{0.769} \pm \textbf{0.006}$ & $\textbf{0.771} \pm \textbf{0.006}$\\
      & & \checkmark & & $0.806 \pm 0.006$ & $0.778 \pm 0.007$ & $0.763 \pm 0.008$ & $0.770 \pm 0.007$ & $0.828 \pm 0.007$ & $0.808 \pm 0.010$ & $0.793 \pm 0.009$ & $0.800 \pm 0.009$\\
      & & & \checkmark & $0.784 \pm 0.009$ & $0.756 \pm 0.010$ & $0.742 \pm 0.011$ & $0.749 \pm 0.010$ & $0.808 \pm 0.010$ & $0.788 \pm 0.013$ & $0.773 \pm 0.012$ & $0.780 \pm 0.012$\\
    \hline \hline
    \end{tabular}
    }}
    \label{tab:ablation_study_phi}
\end{table*}

\subsubsection{Impact of Backbone Depth}
\cl{
To assess the influence of different backbone depths on the model's performance, we chose three variants of ResNet architecture - ResNet18, ResNet34, and ResNet50 - for our experiments. The corresponding results are presented in Table \ref{tab:ablation_study_depth}.
Surprisingly, the shallowest architecture, ResNet18, outperforms its deeper counterparts on both datasets as shown in Table \ref{tab:ablation_study_depth}. This suggests that a more complex model does not guarantee better performance, highlighting the importance of model depth selection for balancing accuracy and computational efficiency.
}

\begin{table*}[htp!]
    \caption{
    Ablation study of backbone depth.
    }
    \centering
    \cl{
    \scalebox{0.9}{
    \begin{tabular}{c | c c c c|  c c c c}
    \hline \hline
    \multicolumn{1}{c|}{Backbone} & \multicolumn{4}{c|}{PTB-XL} &  \multicolumn{4}{c}{CPSC2018}\\
      & Accuracy & Precision & Recall & F1-score & Accuracy & Precision & Recall & F1-score\\
    \hline
      ResNet18 & $\textbf{0.835} \pm \textbf{0.003}$ & $\textbf{0.804} \pm \textbf{0.005}$ & $\textbf{0.787} \pm \textbf{0.006}$ & $\textbf{0.792} \pm \textbf{0.005}$ & $\textbf{0.859} \pm \textbf{0.003}$ & $\textbf{0.838} \pm \textbf{0.012}$ & $\textbf{0.781} \pm \textbf{0.003}$ & $\textbf{0.782} \pm \textbf{0.003}$\\
      ResNet34 & $0.821 \pm 0.007$ & ${0.801} \pm {0.009}$ & ${0.784} \pm {0.004}$ & ${0.789} \pm {0.003}$ & ${0.854} \pm {0.012}$ & ${0.832} \pm {0.008}$ & ${0.776} \pm {0.007}$ & ${0.780} \pm {0.008}$\\
      ResNet50 & ${0.818} \pm {0.004}$ & ${0.798} \pm {0.005}$ & ${0.785} \pm {0.013}$ & ${0.788} \pm {0.011}$ & ${0.851} \pm {0.009}$ & ${0.833} \pm {0.006}$ & ${0.774} \pm {0.010}$ & ${0.778} \pm {0.005}$\\
    \hline \hline
    \end{tabular}
    }}
    \label{tab:ablation_study_depth}
\end{table*}

\section{Conclusion and Future Work}
\label{conclusion}
\cl{
The SCDNN proposed in this study effectively fuses information from the time and spectral domains, thanks to a key architectural innovation - SATSE block. This block, equipped with trainable thresholds, adeptly filters high- and low-frequency domain information.
SCDNN's efficacy is rigorously tested on two public, large-scale, 12-lead ECG signal databases covering diverse classification tasks. In comparison to existing state-of-the-art methodologies, our model shows a significant advantage across all metrics, underscoring the utility of spectral domain knowledge for minimizing convolutional layer information loss. Importantly, the SATSE's ability to adaptively select appropriate frequency modes in the spectral domain demonstrates a tangible benefit. The crucial contribution of the SATSE is further substantiated by our comprehensive ablation studies.
The SCDNN framework's applicability extends beyond its current implementation. Its potential use in various fields like computer vision, natural language processing, speech signals, and physics systems will be further investigated. Notably, SCDNN boasts the shortest inference time, the highest performance on all metrics for the two datasets, and the lowest standard deviation values compared to other models, underscoring its impressive efficiency and effectiveness.
}

\section*{Acronyms}

\begin{acronym}[AAAAA]
\acro{NN}{Neural Network}
\acro{AUC}{Area under Curve}
\acro{ConvF}{Convolutional Fourier}
\acro{DNN}{Deep Neural Network}
\acro{FFT}{Fast Fourier Transformation}
\acro{ACC}{Accuracy}
\acro{FNO}{Fourier Neural Operator}
\acro{SCDNN}{Spectral Cross-domain Neural Network}
\acro{BLUE}{Best Linear Unbiased Estimator}
\acro{3D-Var}{3D Variational}
\acro{RNN}{Recurrent Neural Network}
\acro{IFFT}{Inverse Fast Fourier Transformation}
\acro{CNN}{Convolutional Neural Network}
\acro{LSTM}{Long Short-term Memory}
\acro{SATSE}{Soft-adaptive Threshold Spectral Enhancement}
\acro{ECG}{Electrocardiography}
\acro{GBDT}{Gradient Boosting Decision Tree}
\acro{SVM}{Support Vector Machine}
\acro{RF}{Random Forest}
\end{acronym}

\section*{Main Notations}
\begin{table}[ht!]
    \centering
    \cl{\begin{tabular}{ p{3.5cm} p{15cm}}
$f^{(j)}_{i,k}$ & output of the Res block in SATSE\\
$C_i$ & number of channels in the $i^\textrm{th}$ SATSE block\\
$L_i$ & signal length in the $i^\textrm{th}$ SATSE block\\
$\mathcal{F}, \mathcal{F}^{-1}$ & FFT and IFFT operators\\
$\sigma^{L}(.), \sigma^{H}(.)$ & sigmoid functions for high- and low-frequency
\\
$\gamma_i$ & slope of the modified sigmoid function\\
$\varphi_{i}$ & trainable threshold ratio\\
$f^{S,(j)}_{i,k}$ & spectral domain features\\
$\widetilde{f^{S^H}_{i,k}}$ & filtered high/low-frequency features\\
$\{ f^{H'}_{i,k}\}, \{ f^{L'}_{i,k}\}$  & inversed Fourier sequences \\
$W_{i,k}$ & trainable weight matrices in SATSE
blocks \\
$O_{i}^{\ac{SATSE}}$ & output of the SATSE block \\
    \end{tabular}}
\end{table}

\section*{Acknowledgement}

Sibo Cheng and Rossella Arcucci acknowledges the support of the Leverhulme Centre for Wildfires, Environment and Society through the Leverhulme Trust, grant number RC-2018-023 and the EP/T000414/1 PREdictive Modelling with
Quantification of UncERtainty for MultiphasE Systems (PREMIERE). Weiping Ding acknowledges the National Natural Science Foundation of China under Grant 61976120, the Natural Science Foundation of Jiangsu Province under Grant BK20231337 and the Natural Science Key Foundation of Jiangsu Education Department under Grant 21KJA510004.

\bibliographystyle{IEEEtran}
\bibliography{IEEEabrv,Bibliography}

\vfill


\end{document}